# Ethnography and Machine Learning: Synergies and New Directions
**[PRE-PRINT VERSION]**

Zhuofan Li and Corey M. Abramson[1]

**Abstract**
Ethnography (social scientific methods that illuminate how people understand, navigate and shape the real world contexts in which they live their lives) and machine learning (computational techniques that use big data and statistical learning models to perform quantifiable tasks) are each core to contemporary social science. Yet these tools have remained largely separate in practice. This chapter draws on a growing body of scholarship that argues that ethnography and machine learning can be usefully combined, particularly for large comparative studies. Specifically, this paper (a) explains the value (and challenges) of using machine learning alongside qualitative field research for certain types of projects, (b) discusses recent methodological trends to this effect, (c) provides examples that illustrate workflow drawn from several large projects, and (d) concludes with a roadmap for enabling productive coevolution of field methods and machine learning.

## Introduction[2]

Ethnography and machine learning (ML) are core tools for contemporary social science, yet they remain largely separate in practice. On a surface level, this seems logical. Ethnography is the term applied to a family of social scientific methods that examine how people understand, navigate and shape the contexts in which they live out their lives. Classical and contemporary examples of ethnography include examinations of urban neighborhoods (Clerge, 2019; Sánchez-Jankowski, 2008; Whyte, 2012), schools (Lareau, 2003; Willis, 1981), workplaces (Burawoy, 1979; Sallaz, 2019), hospitals (Glaser & Strauss, 1965; Reich, 2014), and myriad other contexts. ML refers to computational techniques that leverage big data and statistical learning models to identify patterns and improve performance on quantifiable tasks such as classifying documents and identifying topics (Garip & Macy, in this volume; Grimmer et al., 2021, 2022; Jordan & Mitchell, 2015; Molina & Garip, 2019).

Yet a growing body of scholarship argues that ethnographic methods and ML can be usefully combined. A growing body of research suggests that combining methods with ML can offer substantial benefits. (Brandt, in this volume; Brooker, 2022; Li et al., 2021; Nelson, 2020; Pardo-Guerra & Pahwa, 2022; Santana and Nelson, in this volume). This is timely be- cause of a number of factors: (a) the growing toolkit for integrating computational social science (CSS) and qualitative analyses of texts (Grimmer et al., 2022), (b) the expansion of team-based ethnographic research that integrates computation (Abramson et al., 2018, Edin et al. 2024), and (c) the growth of qualitative data repositories that can benefit from combined methods (Murphy et al., 2021). Furthermore, there are core synergies at the intersection of methodological projects that aim to "scale up" field methods to identify patterns (Abramson & Dohan, 2015; Bernstein & Dohan, 2020) and "scale down" big data to be more precise and contextual (Breiger, 2015; Lazer et al., 2014; McFarland & McFarland, 2015).
This chapter addresses recent advances in using ML alongside ethnography and illustrates how researchers can use them in their own research. We begin by reviewing core background concepts and debates. Then we move on to the value and challenges of combining ethnography and ML. Next, we provide a sample workflow drawn from our implementation across several large projects. We conclude with a roadmap for facilitating productive exchange between ML and field methods.

---

[1] Zhoufan Li Ph.D. is an assistant professor of Sociology at Virginia Tech. Corey M. Abramson Ph.D. is associate professor of Sociology at Rice University, and PI of the Computational Ethnography Lab.
Email: corey.abramson@rice.edu.
[2] We thank Dan Dohan and the members of the University of California San Francisco Medical Cultures Lab—Sarah Garrett, Alma Hernandez, and Melissa Ma, in particular—for generously allowing us to report methodological work from our collaborative projects, and Ron Breiger for his advice and encouragement along the way. We would like to acknowledge the following sources of funding which enabled this work: NIH DP1AG069809 (PI: Dohan), NIH R01CA152195 (PI: Dohan), and a Research, Discover, and Innovation Faculty Seed Grant at the University of Arizona (PI: Abramson). The opinions are those of the authors not funders, and any errors are our own.

## Ethnographic Research in the Era of "Big Data"

Ethnographic research typically involves observing people's actions, speech, and interactions within real-world settings, a method known as participant observation, for a process called fieldwork (Becker, 1958). In recent decades, ethnographic practices have expanded to include in-depth qualitative interviews, analyses of historical documents, digital interactions, and computationally aided comparisons.1 For this chapter, we use the term ethnography in an inclusive and pragmatic way, focusing on the following shared practical challenges that exist across traditions (and that may benefit from ML): (a) making sense of semi-structured Ethnographers have responded to the expansion of computational methods and big text reflecting the speech and behaviors of human subjects and (b) doing so in a way that maintains interpretive depth and avoids relying solely on quantitative reduction.

Ethnographers have responded to the expansion of computational methods and big data in varied ways (Albris et al., 2021; Bjerre-Nielsen & Glavind, 2022; Brooker, 2022). Some have used a traditional form of humanist ethnography as critique of technological trends (Knox & Nafus, 2018). Others caution against an over-reliance on quantitative metrics, highlighting ethnography's value as a complementary, case-based alternative to variable-driven social science (Grigoropoulou & Small, 2022; Small, 2009). Some argue that the development of computational tools enables increased scalability and generalizability that may be useful for certain types of projects (Abramson et al., 2018; Bernstein & Dohan, 2020; DeLuca, 2022; DeLuca et al., 2016; Edin et al., in press; Murphy et al., 2021). Scholars have argued that data analysis soft- ware can streamline existing qualitative procedures, even in traditional approaches that rely on iteration and inductive analyses of text patterns (Deterding & Waters, 2021; Dohan & Sánchez-Jankowski, 1998; Friese, 2019).

## New Approaches

Two new avenues of inquiry use computers to go beyond traditional ethnography. The first trend has been labeled digital ethnography. Previously, researchers had used participant observation to examine online communities, interactions, and games in ways that transposed traditional anthropological and sociological field methods (Boellstorff, 2015; Boellstorff et al., 2012; Hine, 2011; Wilson & Peterson, 2002). As human experience increasingly moves online, digital ethnography has evolved to encompass a wider range of methods for analyzing diverse digital content (Coleman, 2010) such as social media posts, online videos, livestreams (Taylor, 2018), and even computer code (Rosa, 2022). A second trend involves applying tools from CSS to data generated using ethno- graphic methods, including participant observation and in-depth interviews. These approaches are sometimes labeled computational ethnography or machine anthropology and aim to find ways to analyze, visualize, and leverage the large volumes of data generated in contemporary qualitative studies using the growing set of tools at the intersection of qualitative research and computational text analysis (Abramson et al., 2018; Bonikowski & Nelson, 2022; DiMaggio et al., 2013; Grimmer et al., 2022; Mohr, 1998; Nelson, 2020; Pardo-Guerra & Pahwa, 2022). This vein of scholarship connects with a growing cadre of computational social scientists who argue there is value in "scaling down" big data analyses of patterns to look at the qualitative content and argue that the correlate of scaling up ethnography can aid in generating insights that may be obscured by either a macro (big data) or micro (traditional ethnographic) focus alone (Abramson et al., 2018; Breiger, 2015; DiMaggio, 2015; Lazer et al., 2014; McFarland & McFarland, 2015).

## Reasons for Integrating ML Into Field Research

Recent works aim to connect advances in ML to ethnographic approaches—including grounded theory (Nelson, 2020), the extended case method (Pardo-Guerra & Pahwa, 2022), and post-positivist realism (Abramson et al., 2018). Table 14.1 offers a summary of various ways ML and other formal and computational analyses have been applied to text data across a variety of research paradigms. These formal and computational methods have broadly facilitated the identification of new insights that might be missed in conventional analysis, the discovery of new opportunities to advance theory, the visualization and sharing of patterns that are subsumed in narratives, the integration of data types in scaled team projects, and advances in addressing concerns about the scale, openness, and validation of ethnographic findings. The table is not meant to be exhaustive but rather to show how these fields have historically overlapped.

ML and ethnography have additional synergies around the overlap between classic ethnographic and computation concerns with iteration, scale, and context. We illustrate how these overlaps are related to emergent approaches in Figure 14.1. First, both ML and ethnography are iterative, as they involve cycling through possible classifications of data to learn patterns, adjust expectations, and ensure that a resulting explanation maps on to new observations and emergent insights in the field.

**[TABLE 14.1] ComutationalEthnography and Approaches to Qualitative Data Analysis**

| | Quantitative Analysis of Text | | Computational Social Science | | | Traditional Qualitative Sociology | |
|---|---|---|---|---|---|---|---|
| | **Dictionary-based Statistical Analysis of Text** | **Rule-based Human Coding** | **Unsupervised Machine Learning** | **Supervised Machine Learning** | **Formal Analysis of Codes** | **Semi-Structured Content Analysis** | **Grounded Theory** |
| **Representations** | Word Frequency | Human | Word Frequency or Co-occurrence | Word Frequency or Word Embeddings | Code Co-occurrence | Human | Human |
| **Classifiers** | Human | | Clustering<br>Topic Modeling<br>Community Detection | Logistic Regression<br>Support Vector Machine<br>Neural Network | Log-linear Model<br>Multiple Correspondence Analysis<br>Bayesian Rule Set | | |
| **Data Requirement** | Varies | Low | Medium | High | Low | Varies | |
| **Evaluation** | Inter-Reader Reliability or Interpretability | | Interpretability | Predictive Accuracy | Goodness-of-fit<br>Predictive Accuracy<br>Bayes Factor | Inter-Reader Reliability or Interpretability | |
| **Reproducibility** | Very Low | Low | High | High | High | Low | Very Low |
| **Coherence** | | | | Depends | | | |
| **Common Sense** | Too Little | Low | Low | High | Low | High | Too Much |
| **Scalability** | Very Low | Low | Very High | High | Very High | Low | Very Low |
| **Role in Computational Ethnography** | As needed | Combined with SML and FAC | Before Human Coding | With Partial Human Coding | After Coding | Combined with SML and FAC | As needed |

**[FIGURE 14.1] Venn Diagram of Overlaps Between Ethnography and ML**

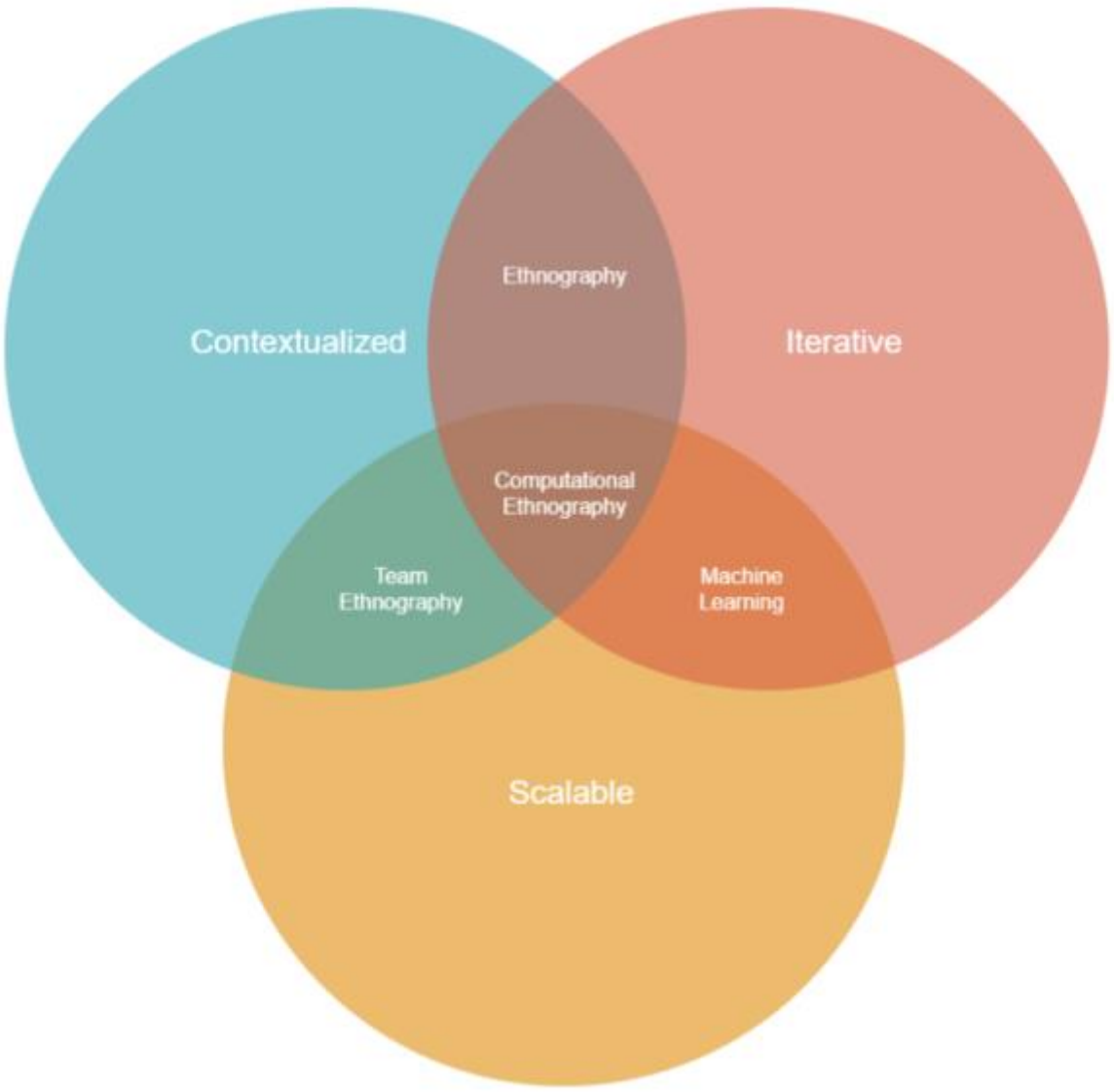

Second, both ethnography and ML engage with social scientific concerns with scale. ML offers powerful tools to scale human input for analyzing new data. However, it is agnostic about the underlying empirical mechanisms, often borrowing from whatever improves its performance, including irrelevant and prejudiced information, in ways that raise concerns about "ecological validity" (Cicourel, 1982) and "algorithmic biases" (Barocas et al, 2019; Danks & London, 2017; Hajian et al., 2016; Hanna et al., 2020). Ethnography provides a way of providing evidence that is deeply grounded in observations of real-world settings, but it has had challenges scaling and has been perennially critiqued for lacking generalizability and transparency (Goldthorpe, 2000; Lubet, 2018). Team projects aim to address sampling challenges for those in realist traditions (Abramson et al., 2018; DeLuca et al., 2016; Edin et al., in press), but they also present issues of interpretation and concerns with rigidity. Technical solutions for efficiently scaling core tasks, such as coding, are sparse and may benefit from the tools of ML (Li et al., 2021; Nelson et al., 2021).

Finally, both ML and ethnography engage with issues of context. Ethnography places context at its core, examining the extent to which data are grounded in a neighborhood, institution population, or cases(s) (Gong & Abramson, 2020). ML has increasingly accounted for semantic context through the examination of word positions in text segments, but it still grapples with contextualization on a conceptual, if not technical, level (Arseniev- Koehler, 2022).

In sum, ML has a core grounding in iteration and strength in scaling, but it is still an underdeveloped toolkit for making sense of social context. Ethnography, likewise, has a core grounding in iteration. In contrast to ML, it has deeply developed resources for understanding social context, but it has underdeveloped tools for addressing is- sues of scale. Approaches like computational ethnography aim to leverage both traditional field research and ML to produce forms of inquiry that might combine some of the strengths and address some of the weaknesses of each approach in isolation. This combination is consistent with the methodological pluralism that contributes vibrancy to qualitative sociology (Lamont & Swidler, 2014), ethnography (Abramson & Gong, 2020), and hybrid approaches to text analysis (Nelson, 2020; Pardo-Guerra & Pahwa, 2022).

## A Pragmatic Approach to Integrating ML and Ethnography

In this chapter, we take a pragmatic approach. Instead of integrating ML as a tool tied to a specific epistemic or methodological tradition, we argue that ML can aid in many practical tasks that are essential to the work of ethnographers across traditions. While traditions vary in how they approach these tasks (Gong & Abramson, 2020), they are common in most contemporary veins of social scientific and policy research (Abramson, 2021). These core practical tasks may include the following:

- *Research design:* deciding on a question or topic, selecting site(s), gaining access, developing an interview schedule, securing funding, submitting to institutional reviews.
- *Observing people and places*: spending time with people in settings ranging from street corners to homes, bars, and online spaces.
- *Writing field notes*: turning jottings from observations into records that can be analyzed.
- *Interviewing subjects*: conducting unstructured or semi-structured interviews with subjects.
- *Transcribing interviews:* turning audio recordings into text that can be analyzed later.
- *Exploring patterns:* identifying and charting patterns or themes, seeing how they vary over time, between people, or across contexts.
- *Coding text:* indexing field notes and interview segments as instances of themes, categories, concepts, or variables of interest.
- *Writing memos:* writing analytical notes for oneself or one's team to make sense of experiences, patterns, and recorded data.
- *Representing findings:* writing to convey to a broader audience or public an understanding of the subject, site, and people involved in a study in a way that would not be possible using different methods.
- *Sharing data:* creating open-science repositories for the sharing and reuse of anonymized ethnographic data.

## Using ML in an Ethnographic Research Workflow

We now turn to a sample workflow that illustrates where and how ML can aid in core tasks of ethnographic studies, with a focus on scaling human coding and visualizing patterns of data (Figure 14.5). To illustrate our workflow, we use the Patient Deliberation Study, a mixed-method, multi-site ethnographic dataset that consists of a total of 196 in-depth interviews with advanced cancer patients, caregivers, and healthcare providers and 12,000 pages of observational field notes, collected for a comparative ethnographic study of experiences of metastatic cancer (Abramson & Dohan, 2015). We use ATLAS.ti version 22, Python 3.8, Jupyter Notebook, and Microsoft Excel.

Here, we focus on core arenas in which ML has broad utility for field studies: (a) data collection, (b) data contextualization and exploration, (c) data visualization, and (d) coding.

## Data Collection

### *Writing Field Notes*

Ethnographers increasingly use a variety of digital tools to help them take field notes: writing software such as Scrivener, Ulysses, or iA Writer (Jackson, 2015), and qualitative data analysis (QDA) software such as ATLAS.ti, NVIVIO, or MAXQDA to manage data from field research (Dohan & Sánchez-Jankowski, 1998; Friese, 2019). Some have experimented with broadcasting their field notes via social media, whereby ML- based recommendation algorithms determine who gets to see the content, although this creates issues with sensitive human subjects data (Wang, 2012). Most recently, researchers and companies have started to commercialize ML systems that can generate summaries, answer questions, and even write creatively based on scientific literature as input (Hutson, 2021; Van Noorden, 2022). The most exciting example is ChatGPT by OpenAI, the latest iteration of a conversational chatbot driven by a large language model (LLM) that uses deep learning algorithms and massive human-generated text training data to mimic the human ability to write. ChatGPT allows users to interact with a robot assistant that can answer questions, write prose, and even code computer programs following user instructions.

There has been heated debate over whether ChatGPT should be recognized as a legitimate tool for academic writing and even research itself. Researchers have found that ChatGPT can write scientific abstracts and that professional reviewers had significant difficulty distinguishing such abstracts from human-written abstracts, the former of which was even more readable and succinct (Gao et al., 2022). PNAS, for example, has required that ChatGPT be acknowledged in the materials or methods sections of published research but never as an author.2 Yet an even more existential question for ethnographers in the age of AI is whether ChatGPT could change the nature of the method all together: Can ethnographers use AI to help them write field notes and still call their work ethnography? How different are an ethnographer's notes about what they see and an AI system's textual description of video? Should AI generate codes and memos or is this dangerous (Abramson 2023)? Whether and how these ML-powered writing tools will help or hinder ethnography, or transform the meaning of field notes altogether, will be a question for ethnographers to answer in the near future.

*Transcribing Interviews*

Although transcription is often left out of methodological discussions in the published work of sociology as a behind-the-scenes task alongside other seemingly mundane topics like data cleaning, it involves methodological assumptions (audio can be represented as text) and the use of computers (Oliver et al., 2005). ML-powered transcription software services, such as Rev, Otter.ai, Zoom's audio transcription function, and Google Recorder, are increasingly used in interview studies to turn recorded speech into text at lower cost than that of human transcription services. Da Silva (2021) found that Google Recorder provided "good enough" first-pass transcription on interview recordings while reducing the time spent transcribing from 5–8 hours to 1–3.5 hours per interview. The fact that these ML-powered transcription services are mostly cloud-based raises confidentiality and privacy concerns. Notably, some companies claim rights over user-uploaded data and reuse the information for their algorithms in undisclosed ways. Still, the time savings and low cost may lead to the increased use of these services, particularly among those without funded research programs. Some companies have started to offer privacy-aware, on-premises deployment of their ML-powered transcription services, but enterprise-level hardware and licenses are often required. Researchers have also shown that state-of-the-art, off-the-shelf automated speech recognition systems are riddled with racial biases; for example, they may be likely to overrepresent White speakers and underrepresent minority speakers in the training data. English speech by African American speakers is transcribed with a higher average word error rate (0.35) compared to speeches by White speakers (0.19) (Koenecke et al., 2020). Knowledge and reflection about algorithmic biases are therefore important when ethnographers utilize these ML-powered speech-to-text transcription services.

*Data Management*

**[FIGURE 14.2] Example of ML-Compatible Ethnographic Data File**

| Name | Date modified | Type | Size |
|---|---|---|---|
| 4020_20110408_DD.txt | 1/26/2022 9:44 AM | Text Document | 27 KB |
| 4021_20110602_SM.txt | 1/26/2022 9:45 AM | Text Document | 53 KB |
| 4022_20110721_SM.txt | 1/26/2022 9:46 AM | Text Document | 26 KB |
| 4023_20110513_DD.txt | 1/26/2022 9:46 AM | Text Document | 0 KB |

*Note: File names may consist of a unique identifier of the participant, the date of the interview, and the name of the interviewer. Most importantly, each component should be separated by an underscore.*

Existing computer assisted QDA software (CAQDAS) has allowed researchers to aggregate, index, code, and retrieve text data for decades (Dohan & Sánchez-Jankowski, 1998; Friese, 2019). Most contemporary commercial programs address early critiques of decontextualization and rigidity by allowing the retrieval of text segments in the context of a longer field record or interview transcript (Deterding & Waters, 2021). However, the optimization of these programs can create technical and methodological challenges with large data sets. Ethnographers would benefit from a flexible integration of CAQDA software with ML workflows.

For optimal integration with an ML workflow for those not using QDA software, we recommend that text be stored in plain text or other easily convertible file types using a consistent data management system. Ideally, transcripts, field notes, documents, and other text data should be stored as .txt files using a consistent naming and formatting structure pre-analysis, so that these .txt files can be stored in a subdirectory, as shown in Figure 14.2. This allows simple Python scripts to aggregate data into either a one- paragraph-per-line or a one-document-per-line format, depending on the desired unit of analysis, data type, location, and data to be extracted as metadata. This one- paragraph-per-line format then allows any text data to be imported into popular programming environments such as Python and R or other ML software for accessing a wide range of ML techniques.

Alternatively, ethnographers can also use off-the-shelf CAQDA software that allows users to store text in a more conventional format and then export the dataset in a one-paragraph-per-line format. Programs like ATLAS.ti and MAXQDA, for ex- ample, allow ethnographers to import, store, retrieve, code, and cross-reference interview transcripts, field notes, documents, and other non-text data types in a single, unified database (Deterding & Waters, 2021) and to export all or a subset of the text data to an Excel spreadsheet that is ML compatible (for a step-by-step tutorial, see Abramson, 2022).

Eventually, the one-paragraph-per-line spreadsheet file should be formatted as follows:

1. Each row should contain a single unit of text (usually a paragraph, a question response, or a document) under a clearly labeled column. In an ATLAS.ti-generated spreadsheet, the text would appear in the "Quotation Content" column.
2. Each row should be indexed by a document-level identifier that identifies the source of the text in another clearly labeled column (e.g., the "Document" column in an ATLAS.ti-generated spreadsheet).
3. Each row should be indexed by another paragraph-level identifier that allows for the reconstruction of the original document in chronological order (e.g., the "Reference" column).
4. Each row should also contain another column that lists any human-generated codes. In an ATLAS.ti-generated spreadsheet, multiple codes will be separated by a new line inside a cell in the "Codes" column. Other separators such as a comma or colon may be used if the codes themselves contain no such punctuation.
5. Each row may contain additional columns that include metadata about the unit of text, such as the name of the speaker, the time of the interview, or the location of the field notes.

Figure 14.3 provides an example of how data formatted in this way would appear.

**[FIGURE 14.3] Example of Ethnographic Data in One-Paragraph-per-Line Format, Using Publicly Available, Non-Human-Subject Data**

| | C | D | E | F | G |
|---|---|---|---|---|---|
| 1 | **Document** | **Section** | **Speaker** | **Codes** | **Quotation Content** |
| 2 | Henry_B._Abajian | Educational Background | Nebeker | Background | This is an interview with Henry Abajian on the |
| 3 | Henry_B._Abajian | Educational Background | Abajian | Background | In 1938 I graduated with an electrical engineer |
| 4 | Henry_B._Abajian | Educational Background | Nebeker | Background | What were you particularly interested in at tha |
| 5 | Henry_B._Abajian | Radiation Lab | Abajian | | It was all power engineering, and the electroni |
| 6 | Henry_B._Abajian | Radiation Lab | Nebeker | | That first year of operation. |
| 7 | Henry_B._Abajian | Radiation Lab | Abajian | | Yes. How I got there was interesting. The heac |
| 8 | Henry_B._Abajian | Radiation Lab | Nebeker | | In those days radar was highly classified. Did y |
| 9 | Henry_B._Abajian | Radiation Lab | Abajian | | No. Not a bit. Wes Hall didn't know too much |
| 10 | Henry_B._Abajian | Radiation Lab | Nebeker | | Did you arrive and then get briefed in what wa |
| 11 | Henry_B._Abajian | Radiation Lab | Abajian | | After I got in was when I first learned about it. |

Note: The columns should include row ID, interview or field note ID, any metadata (section name, speaker name), any analytical codes that ethnographers have applied, and the full text content. In this example, each paragraph is defined by turn taking in the interview, and we were interested in renowned scientists' and engineers' narrative about their family and educational backgrounds.

**Contextualization and Exploration**

Case and field-site selection is a crucial starting step of ethnographic research for a broad range of ethnographic traditions (Abramson & Gong, 2020; Timmermans & Tavory, 2012). Traditional accounts of ethnographic approaches emphasize casing as theoretical, yet as Pardo-Guerra and Pahwa's (2022) comparison between ethnography and CSS alludes to, casing is more than a theoretical moment (Lichterman & Reed, 2015). Ethnographers often build up background knowledge about potential cases and field sites to make an informed decision, which involves existing literature, news articles, documents, social media posts, and other secondary materials. Ethnographers may explore these supplementary data using CAQDA and/or ML techniques in a more content-analysis-style step before diving into fieldwork—for instance, by following the computational grounded theory framework proposed by Nelson (2020) or the extended computational case method framework proposed by Pardo-Guerra and Pahwa (2022).

*Topic Modeling*

A practice shared by computational content analysis frameworks that may be useful here is the deployment of topic modeling, an ML method that uses word frequencies in documents to determine distinct thematic clusters of words (Blei et al., 2003; DiMaggio et al., 2013) in order to discover "topics" in the corpus of interest. These topics then help researchers track semantic changes over time (Fligstein et al., 2017), define an initial set of analytical themes and codes (Nelson, 2020), and delineate the boundaries of the case (Pardo-Guerra & Pahwa, 2022). Although not yet used extensively in ethnography, these topics can potentially provide more discursive contexts alongside the historical accounts seen in projects like those described in Venkatesh (2009), Gong (2020), and Lara-Millán (2021). Table 14.1 in Nelson (2020) lists popular packages for off-the-shelf topic modeling in Python and R. Ethnographers can also run standalone topic modeling applications without using programming environments through the web-based jsLDA by David Mimno (n.d.) or the Windows-based ConText by a team led by Jana Diesner (2020). The fact that ConText requires one-document-per-file .txt files whereas jsLDA requires one- paragraph-per-line .csv files attests to the importance of formatting that we discussed in the previous section.

Topic models can help identify recurrent themes and topics in news articles and social media posts, but caution is advised in ensuring that interpretation is contextualized in ways that map onto a qualitative reading of the content. This is because most topic models make strong and, in many cases, unrealistic statistical assumptions about the text being modeled, including that (a) word order does not change the meaning of the text, (b) topics in each document follow a unimodal, Dirichlet distribution, and (c) the number of topics to be discovered can be specified a priori. These assumptions make topic models suitable for analyzing documents that (a) use domain-specific and highly precise vocabulary (as opposed to many unspecified verbs, pronouns, filler words, and other words common in oral expressions), (b) focus on one distinct topic (as opposed to going back and forth between multiple topics), and (c) do not involve extended, sophisticated narratives (as opposed to, for example, legal arguments). Even in these documents, the risk is imputing typologies that are statistical artifacts rather than sociological realities, and the ethnographic concern with contextual understanding is even more important when explaining the accounts and actions of human subjects.

Another crucial consideration before applying topic models is the preprocessing of the text. Most preprocessing steps, including tokenization and stemming, are already automated in popular natural language processing (NLP) packages listed by Nelson (2020), such as NLTK, openNLP, and tm. However, the identification and removal of stop words may require more hand curation by ethnographers. Stop words are words such as "the," "a," "by," "for," and "and," which are exceptionally common in English but carry little semantic information relative to the noise they add. Because topic models rely heavily on word frequencies, removing stop words is critical to the performance and interpretability of topic models. Lists of common stop words are typically provided by popular NLP packages, but often researchers customize those lists and manually re- move significant stop words in their own text.

*Word Embeddings*

Word embedding models like Word2vec are another type of ML model of text data that encode the meaning of a word in a high-dimensional vector such that words that tend to be used in similar contexts are represented by vectors that are closer to each other in the vector space. To encode the meaning in this way, a neural network model is trained by predicting the target word given other words in its preceding and subsequent contexts or by predicting the context words given the target word (Le & Mikolov, 2014; Mikolov et al., 2013). In this way, word order, narrative structure, and other higher order linguistic contexts, which topic models have difficulty modeling, can be partly taken into account when word embedding models represent the meaning of a word. These fine-grained representations of semantic relationships among words allow us to explore more dimensions of meaning in ethnographic data.

Word embedding models can be pretrained on large corpora of English language text, such as Google News or Wikipedia, for predictive tasks. But for exploratory purposes, we can also train word embedding models to represent our particular ethnographic data. Because of the tiny size of our dataset by the ML standard, word embeddings will "overfit" our data—memorizing context-specific word usages too well to generalize to other context—but those context-specific word usages are exactly what ethnographers are trying to understand and can provide a useful alternative view of the data for deeper interpretive reading.

Word embeddings can then be combined with other ML techniques to explore the semantic context of the data. For example, word embeddings can detect context- specific (a) synonyms, or words used in similar contexts, (b) antonyms, or words used in contrary contexts, and (c) word analogies, such as "King is to queen as men are to [blank]." We can apply nonparametric clustering algorithms, such as k-mean clustering,
on cosine distances among word embeddings, and visualize the semantic map on a two-dimensional plane using dimensionality reduction techniques such as truncated singular value decomposition and principal component analysis. These explorations of latent semantic relationships help ethnographers connect strings of meanings across multiple observations. We can also obtain document-level vector representations in a similar way. Doc2vec is an extension of word2vec that learns an additional vector for each document or any higher level unit of text on the basis of word vectors learned by word2vec (Le & Mikolov, 2014). These document embeddings allow us to compare se- mantic similarities among paragraphs or among interviews.

**[FIGURE 14.4] Truncated Singular Value Decomposition of a Word Embedding of Cancer Narratives in 69 PtDelib Interviews**

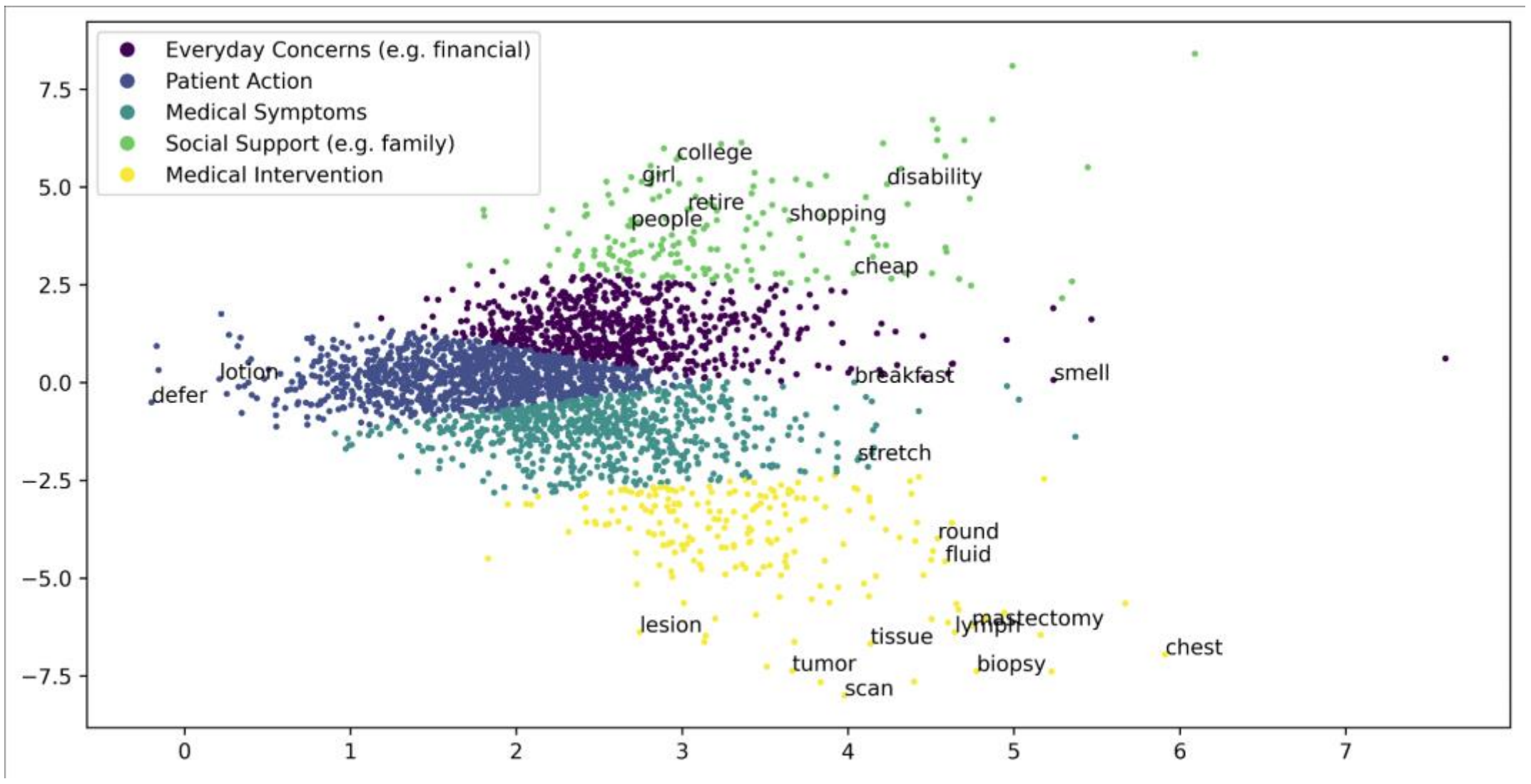

*Note: Words that were used in similar contexts appear close to each other. The vertical axis differentiates between medicalized and everyday experiences. Color indicates semantic clusters*

*identified by a k-mean clustering algorithm. Only 10% of the words on the periphery are labeled for better readability.*

Figure 14.4 presents a truncated singular value decomposition of a word2vec model of all words that interviewees used in the Patient Deliberation Study to describe their experience with cancer, whereby words are positioned against each other depending on the extent to which they were used in similar contexts in the interviews. A k-mean clustering algorithm identifies the five most distinct clusters of words, which we interpreted as “Everyday Concerns” (e.g., financial concerns), “Patient Action,” “Medical Symptoms,” “Social Support” (e.g., family support), and “Medical Intervention.” In other words, word usage differentiates along approximately two main dimensions. Along the vertical axis, we observe a clear distinction between the everyday and medicalized experiences of cancer; along the horizontal axis, we observe a less clear distinction between intervention and coping. This clustering and decomposition of the semantic space corroborate and contextualize a similar coding schema that human researchers developed based on in-depth reading.

For easy implementation of word2vec and doc2vec, we recommend “genism” in Python and “word2vec” in R.

*Semantic Network Analysis*

A semantic network view of ethnographic data presents words as nodes, and frequently co-occurring words as pairs of nodes connected by an edge that is weighted by the frequency of co-occurrence. The sharp visual contrast between the presence and absence of dense edges makes semantic networks useful not only for analysis but also for visualizing narratives that cannot be easily summarized and presented to readers otherwise. A major limitation of semantic network analysis, which may nevertheless be a major opportunity for ethnographers, is that semantic networks require a lot of human curations to construct. Ethnographers will have to decide, based on their research questions and deep understanding of the data, (a) what to include in order to make the network best reflect the narratives, (b) what to exclude in order to make the resulting network human readable, and (c) what constitute frequently co-occurring words. We recommend comparing between the sentence/document-level co-occurrence and the seed-word designs as a starting point. We also recommend considering limiting networks to nouns and adjectives, or to verbs and adverbs, depending on the question.

A sentence/document-level co-occurrence design includes all words used in a sentence/document. The main advantage of a sentence/document-level co-occurrence design is that it helps ethnographers discover words that are central to narratives. Rule et al. (2015), for example, base their semantic network view of more than 200 years of State of the Union speeches on “frequently occurring noun terms, including multiword phrases” (p. 18309) such as “national security,” “local government,” and “fellow citizens,” and they define co-occurrence over documents spanning a particular period of time because they are interested in lexical change from one period to another. However, a sentence/document-level co-occurrence design often results in networks that are too dense to visualize, and it requires extensive pruning of nodes and edges in ways that preserve the core structure of the narratives. A good pruning rule is hard to find and to justify.

While a sentence/document co-occurrence design starts with the context for locating key words, a seed-word design starts with a human-curated list of keywords and “grows” the network by searching the contexts of these keywords for semantic connectivity. Padgett et al. (2020), for example, construct a semantic network of policy debates among Renaissance Florentine political elites by selecting 15 keywords that indicate common topics of discussion and expanding the network iteratively by searching the contexts of these keywords for correlations with other keywords. We can observe, as Padgett et al. (2020) did, how different concepts or themes (which researchers have al- ready identified) are intertwined in the narratives. By restricting the network to a few keywords and their immediate contexts, this type of semantic network analysis tends to produce much more readable visualizations but also much more limited visions of the text.

The disadvantages of these two designs may be mitigated by combining semantic network analysis with ML. One way to reduce the size of the text before we construct se- mantic networks is to keep only nouns, or only name entities—the names of agents such as a person, a country, an object, or an organization—whose semantic relationships are most likely to be of interest to ethnographers. To identify nouns, we use part-of- speech tagging, which classifies a word in a sequence of text as corresponding to a part of speech. To identify name entities, we use name entity recognition, which classifies a word or a multi-word phrase as part of the name of a person (PERSON), an organization (ORG), a geographic entity (GPE), a nonentity location (LOC), an event (EVENT) or a facility (FAC). ML algorithms for part-of-speech tagging can be used to automate this process with near-perfect accuracy. Another way to simplify dense networks is to reduce the networks of nodes to networks of clusters. ML algorithms, such as hierarchical clustering and modularity-based partition, are often used to cluster nodes based on their connectivity to the rest of the network.

ML can also be used to expand the coverage of seed-word networks. Word embedding—a NLP technique that accounts for word position in text—is used first to identify a list of keywords and then to identify synonyms, antonyms, and/or analogies of these keywords under a specific context, which are often as important as human-curated keywords. For example, in a working paper using interview data from the American Voices Project, we combined word2vec with semantic network analysis to identify synonyms of "pain" and distinct narratives about "pain" from interviews and how those narrative types map onto cultural schemas and social identities (Abramson et al., 2023).

We recommend the "igraph" package in Python and in R for constructing, visualizing, and clustering semantic networks. For part-of-speech tagging and name entity recognition, we recommend "spaCy," also available in both Python and R, which provides easy- to-use functions that tag parts of speech and name entities at an average accuracy of 97% and 85%, respectively.

*Clustered Heatmaps*

Heatmaps also provide a way to visualize and explore how codes cooccur and themes overlap in a data set over the course of a study. Existing QDA software such as ATLAS. ti and MAXQDA offer tools for heatmapping and linking to underlying text. Heatmaps represent each respondent as a column, their attributes (from demographic characteristics to semantic themes uncovered in topic modeling) as rows, and associations/co- occurrences between respondents and their attributes as colored blocks (Abramson & Dohan, 2015). Hierarchical clustering is an unsupervised ML method that can then identify clusters of respondents who share similarities in heatmaps. For example, Garrett et al. (2018) used hierarchically clustered heatmaps to identify styles of medical decision- making in cancer patients. Abramson et al. (2018) argue that clustered heatmaps can be used more broadly to reveal patterns, validate typologies, or show code absence for examining counterfactuals.

For a ready-to-use toolkit on structured data, we recommend ggplot2 in R and Seaborn in Python, two open-source packages for statistical software.

**Coding**

Qualitative coding is the process whereby analysts tag qualitative data. This is one arena, in our estimation, in which ML has clear utility even for more traditional examinations of field data. Coding is akin to the application of researcher hashtags to segments of text: They indicate that the text contains something noteworthy, and they index that content with a theme, concept, category, or variable (Deterding & Waters, 2021) in ways that help with search and retrieval (Dohan & Sánchez-Jankowski, 1998). In addition to ethnographic observations, researchers often code interview transcripts, field notes, historical documents, blogs, Reddit posts, or other content that includes text, images, or other data that are indexed according to analyst interpretation. Codes can be applied and used in myriad ways depending on research aims (to identify themes, to capture time variation, to index good quotes, or to reference speaker disposition), but coding does not imply the reduction of a qualitative data set to numbers or variables. Coding is typically iterative, flexible, and nonexclusive, allowing a segment of text to be tagged with multiple codes and for codes to emerge over the course of a study. Most approaches to coding ethnographic data are completed alongside a deeper reading of underlying text and paired with the writing of longer memos.

One of the problems with coding is that it takes a long time while also requiring the flexibility to identify new patterns in data. Prior attempts for large data sets have involved teams of human coders, which is both time intensive and requires continual active work to ensure that codes are being applied in a consistent way across coders. We have argued that a hybrid, human-ML approach to coding can use contemporary NLP techniques that use ML in order to speed up process while maintaining the iterations between an evolving set of codes and human interpretation (Li et al., 2021). ML has applications beyond aiding in coding. Finding ways to apply codes qualitatively to growing qualitative data sets is one of the core practical and methodological challenges of contemporary field methods (Edin et al., in press; Li et al., 2021; Murphy et al., 2021).

*Using Machine Learning to Scale Human Coding*

After exploratory analysis helps researchers organize data and identify concepts of interest, researchers often proceed to code data more systematically. Once data have been preformatted and a preliminary but stable codebook has been created, ML can help quickly scale initial coding to the rest of your ethnographic data. This process can be repeated at various points in a project, as a coding scheme evolves and new data comes in, in a way consistent with contemporary qualitative iteration (Li et al., 2021). For the sake of simplicity, we provide an example that assumes we are at the following point in a project:

1. We have initial research sites, text data, and a relatively stable codebook.
2. We have applied those codes to a sample of field notes or interview transcripts in QDA software or in ML-compatible files.
3. We have exported the coded data into a one-paragraph-per-line Excel file, as described above.
4. (optional) We have imported the spreadsheet into Python or R and conducted exploratory computational contextualization, analysis, and/or visualization, as described above.
5. We still have a considerable amount of text—maybe a few dozen (or even hundred) interviews plus a year's worth of field notes—that need to be coded for analysis and retrieval.

Here, we use an ML workflow to do the following:

6. Train ML classifiers on a subset of the coded data to scale the codes to the remaining data.
7. Evaluate ML coding on a separate subset of the coded data for obtaining estimates of coding reliability.
8. Remove false positives through a review of positive cases.
9. Adjust the training process for better consistency and reliability.
10. Export the data back to a one-paragraph-per-line Excel file.
11. (Optionally) Import the Excel file back into a QDA program like ATLAS.ti, for further qualitative analysis.

Figure 14.5 below summarizes this hybrid workflow.

**[FIGURE 14.5]** Workflow Combining Qualitative Human Coding with ML

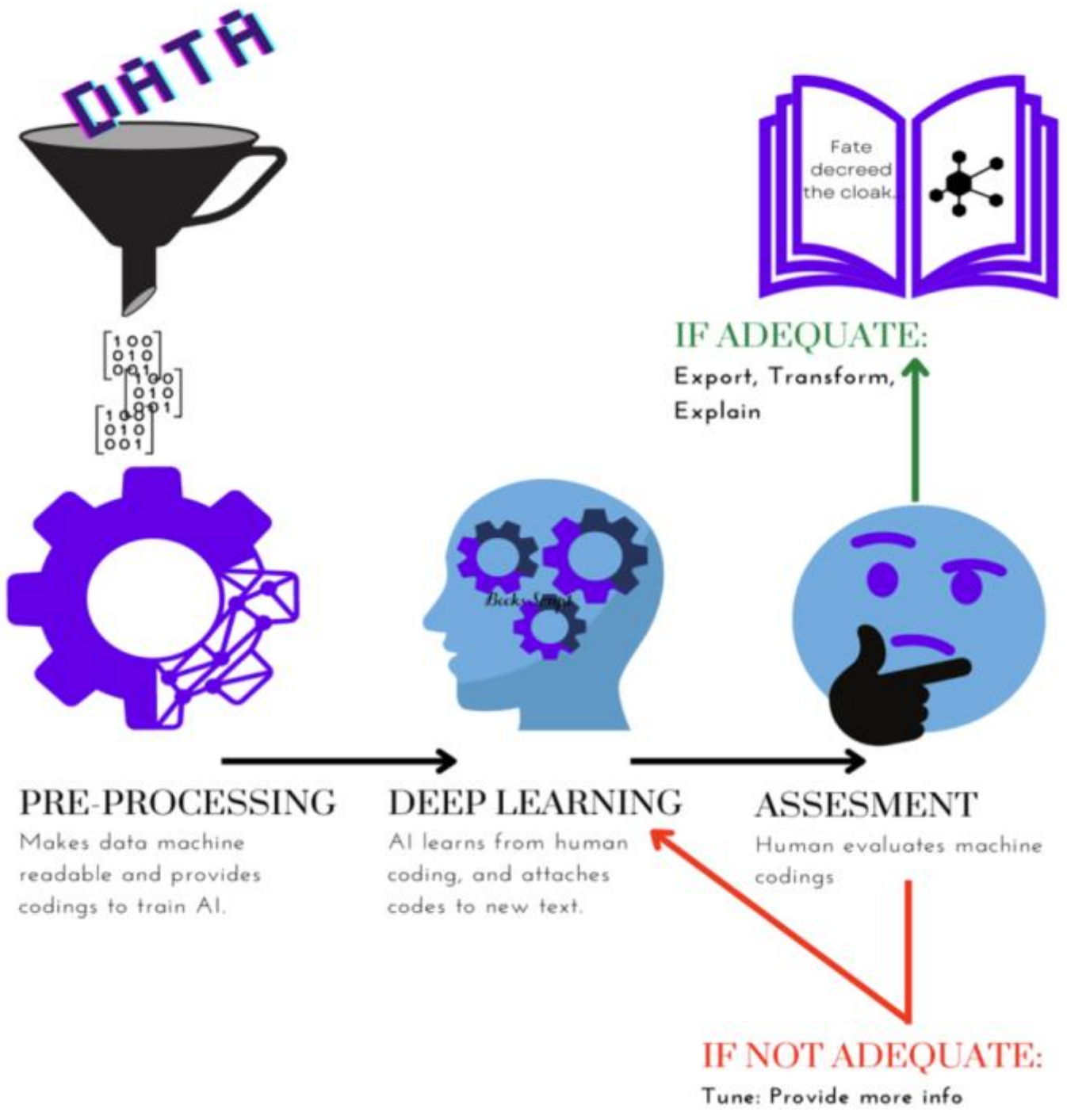

*Performance*

How well does hybrid coding strategy perform on scaling human coding? We find that it works quite well, given our data and aims. We explored these two questions using the Patient Deliberation dataset. A team of ethnographers had coded all interviews and field notes in ATLAS.ti using a modern, iterative coding process (Deterding & Waters, 2021) and achieved an inter-coder agreement, for instance Krippendorff's $\alpha > 0.80$. Our goal was to test whether ML can help us scale human coding on a training sample to the whole dataset. We tested a wide range of ML algorithms on different codes and different sizes of training data.

We reported our preliminary technical results elsewhere (Li et al., 2021), so here we focus on the bigger picture for practitioners in ethnographic research. First, we were interested in how well ML classifiers that are popular in computational content analysis could scale human coding on interview transcripts and field notes, and whether introducing contextualized word embeddings could improve machine performance. Table 14.2 summarizes our results. We tested two ML classifiers that use word frequencies to represent the text and three classifiers that use the pretrained bidirectional encoder representation from transformers (BERT), which is a state-of-the-art, transformer-based, contextualized word embedding model that has revolutionized NLP (Devlin et al., 2018). In word2vec, the same word will be represented by the same vector, but in BERT, the same word in a unique context will have a unique representation. Like word2vec, contextualized word embeddings can also be pretrained on large volumes of text for a general "understanding" of the English language and then fine-tuned on our data for context-specific predictive tasks. BERT, for example, is pretrained on 984 mil- lion words from 11,038 books and another 2,500 million words from English Wikipedia articles. BERT's ability to learn the contextualized representation of language allows it to encode a wide range of linguistic structures—from sub-word features to paragraphs— making it an ideal model for sentence-level language understanding tasks on less structured and less stylized text, such as conversations and narratives. We found that while the number of positive examples and the type of codes (informational versus interpretative) significantly affect the coding reliability of ML algorithms, BERT-based classifiers code more consistently than traditional classifiers do and achieve a usable level of re- liability on informational codes. None of the codes achieved usable reliability on field notes, likely because of their highly unstructured and inconsistent styles.

Second, we were interested in how training BERT-based classifiers on positive training examples could affect machine coding reliability. We suspected that codes that occur more frequently in the data provide more diverse training examples for ML algorithms. We then tested our most efficient algorithm—BERT representations + logistic regression − on another set of codes that are all interpretative but occur in the data at different frequency levels. The second and third rows of Table 14.3 summarize our results. We found that training ML algorithms on more than 400 examples is likely to yield a near-human level of reliability.

Finally, we were interested in how ethnographers could help ML algorithms. We noticed that our ML classifiers tended to achieve a much higher level of recall on our data—a measure of how well a classifier recovers all relevant text, regardless of how many false positives it also recovers. Suppose that you are in a restaurant and order fish and chips, but your server brings you fish, chips, and a bottle of Chardonnay, which you did not order. That server has a recall of 100%, because they recover all the relevant items despite including a false positive, which is straightforward to return. In other words, a classifier with high recall casts a wider net on the data and picks everything up, and a second-pass review allows for the removal of false positives from the smaller set of responses. It is easier for ethnographers to filter out false positives by reviewing responses through rereading a small set of text than it is to identify false negatives that are missed in the entire corpus, so tuning a system for recall and combining with a second-pass review yields accurate results efficiently (Li et al., 2021). The last row of Table 14.3 summarizes our results. We found that combining more than 500 examples, BERT representations, logistic regression classifier, and second-pass human recoding could replicate interpretative human coding on interview transcripts at a near-human reliability with only a fraction of human effort.

**[TABLE 14.3] Machine Coding Reliability by Codes and by Algorithms, 25% Training Data**

| | Medical Test | Social/ Emotional Support | Any Decision Making | Any Patient Information | GPU/TPU Acceleration |
|---|---|---|---|---|---|
| **Type of Code** | Informational | Interpretative | Interpretative | Informational | |
| **Positive Examples (paragraphs)** | 332 | 397 | 824 | 2043 | |
| *Using Word Frequency Representations* | | | | | |
| **Logistic Regression** | 0.41 | 0.34 | 0.39 | 0.58 | |
| **Support Vector Machine** | 0.58 | 0.33 | 0.49 | 0.63 | |
| *Using Contextualized Word Embedding (BERT) Representations* | | | | | |
| **BERT Fine-tuning** | 0.6 | 0.36 | 0.43 | 0.66 | Yes |
| **RoBERT Fine-tuning** | 0.61 | 0.48 | 0.52 | 0.64 | Yes |
| **BERT Representations + Logistic Regression** | 0.56 | 0.41 | 0.48 | 0.69 | Yes |

In sum, we found the following:

1. The introduction of representation learning—in particular, pretrained, contextualized word embeddings such as BERT—significantly improved the reliability of popular ML classifiers across different codes, but it also required a larger amount of training data ( > 400 positive examples and at least the same number of negative examples).
2. Combining ML with a second-pass human recoding of machine coding allowed for an efficient division of labor between ML and human coders and again significantly improved the overall coding reliability.
3. Codes that represent simple language patterns rather than complex ideas that require knowledge of a specific context (e.g., cancer care) can be reliably scaled with fewer examples (because they are easier for pretrained models to identify).
4. The size of training data is a significant predictor of machine performance on coding.
5. After our ML workflow scaled our coding of to all interviews in the dataset, we reimported the coded data back into ATLAS.ti and continued our in-depth qualitative analysis there. Whenever we developed a new code or revised our existing codes, we could easily repeat the workflow anywhere in our analysis.

As coding helps make sense of ethnographic data, alongside memos and reflection, these hybrid procedures that integrate supervised ML into coding workflows offer ethnographers higher level, alternative views of the data that they have just collected, and these procedures quickly establish any revision to the initial codebook or analytical framework.

We implemented BERT using the "PyTorch" (Paszke et al., 2019) and "transformer" (Wolf et al., 2020) libraries in Python. A sample workflow is available via the first author's GitHub repository (Li, 2022) and is readily executable using Google Colab, a free cloud computing service.

**Conclusion**

A growing cadre of researchers has argued for the value of combining traditionally "qualitative" methods, such as ethnography, with contemporary computational tools, such as NLP techniques tied to ML, in order to generate new insights that are at risk of being missed using a single approach in isolation. Researchers have demonstrated potential within a variety of research paradigms connected to ethnography that move be- yond conventional QDA, including post-positivism realism, the extended case method, and grounded theory (Abramson et al., 2018; Nelson, 2020; Pardo-Guerra & Pahwa, 2022). The resulting tools have the potential to advance social science inquiry if used thoughtfully. This chapter has argued that ethnographic research and ML in particular have important overlaps.

While both ethnography and ML are consistent in their use of iteration for under- standing complex phenomena, they differ in how they address issues of context and scale. Rather than seeing one approach as inherently better, we believe both classic ethnography and ML provide synergistic techniques for examining the growing volumes of data seen in contemporary field studies. While a computational ethnography approach that aims to provide new types of analyses is not meant to be a replacement for either big data pattern analysis or micro-level narrative, it may provide a link for understanding middle-range phenomena that connect levels of analysis and/or provide complementary insights. Even if this is not an aim, we have tried to demonstrate how some of the tools of ML can be used to address shared practical problems in ways that can free up time and energy for field researchers.

It is important to acknowledge barriers to the more direct and purposeful combination of ML and ethnographic research that we have suggested for certain types of projects. First, ethnography remains a fragmented field with competing research paradigms and epistemologies that range from positivism to postmodern criticism (Abramson and Gong 2020; Gans, 1999). At best, this can produce meaningful dialogs in which different ways of knowing about the world contribute to rich understandings but can also create confusion and positional struggles about which "ethnographic schools" get to determine the lines between methodological innovation and heresy (Gong and Abramson, 2020). Our focus on the technicality of ethnography and a pragmatic move toward exploiting an emerging toolkit for ethnographers' diverse but interconnected empirical, methodological, and theoretical challenges may provide a way forward. Importantly, a pragmatic agenda also encourages ethnographers to rethink epistemic divisions and invites journals and presses to reconsider how they can integrate novel works that do not fit neatly into preexisting genres and methodological traditions.

Second, the quantitative/qualitative divide may be less entrenched with those trained in the time after mixed methods rose to prominence. In some subfields of sociology, young scholars are expected to have both statistical and qualitative literacies. Yet the number of young scholars who actually receive high quality training in both computational and field methods lags behind the discipline's ambition. In other subfields such as ethnography, a singular approach still holds sway as a gold standard. While this may change as interdisciplinarity grows, there is value in training that integrates computational literacies alongside traditional qualitative competences. This will facilitate the growth of emerging scholars positioned to learn from and engage with the coevolution of field methods and computation.

Third, there are technical barriers to implementing ML and field methods. While computing power can be a bottleneck for large data sets, a reasonably equipped desktop can run ML models to scale human coding using NLP, to visualize clusters of field notes in heatmaps, to generate semantic networks, or to link tables of thematic overlaps to underlying quotations in QDA software. However, learning programming can be time consuming, and software can be expensive, so the development of easy-to-use, open- source tools and courses may help. While the ability of conversational AI systems such as ChatGPT to code based on natural language instructions could help ethnographers circumvent many of these technical steps, uploading confidential data may pose a privacy risk and a violation of current human subject protection regulations.

The integration of ML with ethnography has an additional potential upside beyond what we have described. First, it opens up new possibilities for the growing range of social scientific inquiries and, eventually, an ethnography of algorithms and ML, as ethnographers gain computational literacy and integrate into the computational infra- structure (Lange et al., 2019; Seaver, 2017). Second, the toolkit provided by approaches like computational ethnography can also be applied to other challenges beyond traditional ethnographic practices—such as concerns about data sharing, transparency, and replication in the digital age—by aiding in the cleaning, de-identification, and curation of large data sets (Abramson et al., 2018; Edin et al., in press; Freese & Peterson, 2017; Murphy et al., 2021). Third, the growth of tools for vectorizing and modeling video and image data (Szeliski, 2022), as well as tools for aggregating and composing memos based on data, may open up even more possibilities for new hybrids.

In the end, a broader update of approaches combining ML and field methods will likely depend on expanding qualitative and computational literacy, minimizing technical barriers, and increasing the number of venues for showcasing the scholarship that demonstrates the utility of emergent approaches. Yet we remain optimistic about the coevolution of computational and ethnographic research in part because both ML and field methods provide irreplaceable practical value that benefit from recombination.

## Notes

1. For a fuller treatment of the challenges and potential of this epistemic pluralism, see Abramson and Gong (2020).
2. Questions about the broader uses and ethics of LLMs as they proliferate in industry, science, government, and even qualitative software is a broader question beyond the scope of this paper (Abramson, 2023).

## References

Abramson, C. M. (2021). Ethnographic methods for research on aging: Making use of a funda- mental toolkit for understanding everyday life. In K. F. Ferraro & D. Carr (Eds.), *Handbook of aging and the social sciences* (9th ed., pp. 15–31). Academic Press.

Abramson, C. M. (2022, March 31). Sub-setting qualitative data for machine learning. https://cmabramson.com/resources/f/sub-setting-qualitative-data-for-machine-learning?blogcategory=ATLAS.ti

Abramson, C. M. (2023). A silicone cage: Qualitative research in the era of IA. Medical Culture's Lab Blog. https://www.cultureofmedicine.org/blog/a-silicon-cage-qualitative- research-in-the-era-of-ai

Abramson, C. M., & Dohan, D. (2015). Beyond text: Using arrays to represent and analyze ethnographic data. *Sociological Methodology*, *45*(1), 272–319.

Abramson, C. M., & Gong, N. (2020). Introduction: The promise, pitfalls, and practicalities of comparative ethnography. In C. M. Abramson & N. Gong (Eds.), *Beyond the case: The logics and practices of comparative ethnography* (pp. 1–28). Oxford University Press.

Abramson, C. M., Joslyn, J., Rendle, K. A., Garrett, S. B., & Dohan, D. (2018). The promises of computational ethnography: Improving transparency, replicability, and validity for realist approaches to ethnographic analysis. *Ethnography*, *19*(2), 254–284.

Abramson, C., Li, Z., Prendergast, T., & Sánchez-Jankowski, M. (2023). Inequality in the origins and experience of pain: How people make sense of, and respond to bodily misery. Department of Sociology, University of Arizona.

Albris, K., Otto, E. I., Astrupgaard, S. L., Gregersen, E. M., Jørgensen, L. S., Jørgensen, O., Sandbye, C. R., & Schønning, S. (2021). A view from anthropology: Should anthropologists fear the data machines? *Big Data & Society*, *8*(2), Article 20539517211043656.

Arseniev-Koehler, A. (2022). Theoretical foundations and limits of word embeddings: What types of meaning can they capture? *Sociological Methods & Research*, Article: 00491241221140142.

Barocas, S., Hardt, M., & Narayanan, A. (2019). Fairness and machine learning. https://fairmlb ook.org/

Becker, H. S. (1958). Problems of inference and proof in participant observation. *American Sociological Review*, *23*(6), 652–660.

Bernstein, A., & Dohan, D. (2020). Using computational tools to enhance comparative ethnography. In Abramson & N. Gong (Eds.), *Beyond the case: The logics and practices of comparative ethnography* (pp. 209–237). Oxford University Press.

Bjerre-Nielsen, A., & Glavind, K. L. (2022). Ethnographic data in the age of big data: How to compare and combine. *Big Data & Society*, *9*(1), Article 20539517211069892.

Blei, D. M., Ng, A. Y., & Jordan, M. I. (2003). Latent Dirichlet allocation. *Journal of Machine Learning Research*, *3*, 993–1022.

Boellstorff, T. (2015). *Coming of age in Second Life: An anthropologist explores the virtually human*. Princeton University Press.

Boellstorff, T., Nardi, B., Pearce, C., & Taylor, T. L. (2012). *Ethnography and virtual worlds*. Princeton University Press.

Bonikowski, B., & Nelson, L. K. (2022). From ends to means: The promise of computational text analysis for theoretically driven sociological research. *Sociological Methods & Research*, *51*(4), 1469–1483.

Breiger, R. L. (2015). Scaling down. *Big Data & Society*, *2*(2), Article 2053951715602497. Brooker, P. (2022). Computational ethnography: A view from sociology. *Big Data & Society*, *9*(1), Article 20539517211069892.

Burawoy, M. (1979). *Manufacturing consent: Changes in the labor process under monopoly capitalism*. University of Chicago Press.

Cicourel, A. V. (1982). Interviews, surveys, and the problem of ecological validity. *American Sociologist*, *17*(1), 11–20.

Clerge, O. (2019). *The new noir: Race, identity, and diaspora in Black suburbia*. University of California Press.

Coleman, E. G. (2010). Ethnographic approaches to digital media. *Annual Review of Anthropology*, *39*(1), 487–505.

Da Silva, J. (2021). Producing “good enough” automated transcripts securely: Extending Bokhove and Downey (2018) to address security concerns. *Methodological Innovations*, *14*(1), Article 2059799120987766.

Danks, D., & London, A. J. (2017). Algorithmic bias in autonomous systems. *Proceedings of the 26th International Joint Conference on Artificial Intelligence*, *17*, 4691–4697.

DeLuca, S. (2022). Sample selection matters: Moving toward empirically sound qualitative research. *Sociological Methods & Research*, Article 00491241221140425.

DeLuca, S., Clampet-Lundquist, S., & Edin, K. (2016). Want to improve your qualitative research? Try using representative sampling and working in teams. *Contexts*. https://contexts. org/blog/want-to-improve-your-qualitative-research-try-using-representative-sampling-and-working-in-teams/

Deterding, N. M., & Waters, M. C. (2021). Flexible coding of in-depth interviews: A twenty-first-century approach. *Sociological Methods & Research*, *50*(2), 708–739.

Devlin, J., Chang, M.-W., Lee, K., & Toutanova, K. (2018). BERT: Pre-training of deep bidirectional transformers for language understanding. arXiv. https://doi.org/10.48550/ arXiv.1810.04805

Diesner, J. (2020). Welcome to ConText. https://context.ischool.illinois.edu/index.php DiMaggio, P. (2015). Adapting computational text analysis to social science (and vice versa). *Big Data & Society*, *2*(2), 1–5.

DiMaggio, P., Nag, M., & Blei, D. (2013). Exploiting affinities between topic modeling and the sociological perspective on culture: Application to newspaper coverage of U.S. government arts funding. *Poetics*, *41*(6), 570–606.

Dohan, D., & Sánchez-Jankowski, M. (1998). Using computers to analyze ethnographic field data: Theoretical and practical considerations. *Annual Review of Sociology*, *24*(1), 477–498.

Edin, Kathryn J., Corey D. Fields, David B. Grusky, Jure Leskovec, Marybeth J. Mattingly, Kristen Olson, and Charles Varner. 2024. Listening to the Voices of America. *RSF: The Russell Sage Foundation Journal of the Social Sciences 10*(4),1–31.

Edin, K. J., Fields, C. D., Fisher, J., Grusky, D. B., Leskovec, J., Markus, H. R., Mattingly,M.,Olson, K., & Varner, C. (in press). Who should own data? The case for public qualitative datasets. *RSF: The Russell Sage Foundation Journal of the Social Sciences*.

Fligstein, N., Brundage, J. S., & Schultz, M. (2017). Seeing like the Fed: Culture, cognition, and framing in the failure to anticipate the financial crisis of 2008. *American Sociological Review*, *82*(5), 879–909.

Freese, J., & Peterson, D. (2017). Replication in social science. *Annual Review of Sociology*, *43*(1), 147–165.

Friese, S. (2019). *Qualitative data analysis with ATLAS.Ti*. Sage.

Gans, H. J. (1999). Participant observation in the era of "ethnography." *Journal of Contemporary Ethnography*, *28*(5), 540–548.
Gao, C. A., Howard, F. M., Markov, N. S., Dyer, E. C., Ramesh, S., Luo, Y., & Pearson, A. T. (2022). Comparing scientific abstracts generated by ChatGPT to original abstracts using an artificial intelligence output detector, plagiarism detector, and blinded human reviewers. BioRxiv. https://doi.org/10.1101/2022.12.23.521610
Garrett, S. B., Abramson, C. M., Rendle, K. A., & Dohan, D. (2018). Approaches to decision- making among late-stage melanoma patients: A multifactorial investigation. *Supportive Care in Cancer*, *27*(3), 1059–1070.
Glaser, B. G., & Strauss, A. L. (1965). *Awareness of dying*. Transaction Publishers.
Goldthorpe, J. H. (2000). *On sociology: Numbers, narratives, and the integration of research and theory*. Oxford University Press.
Gong, N. (2020). Seeing like a state athletic commission: Multi-case ethnography and the making of "underground" combat sports. *Ethnography*, *21*(2), 176–197.
Gong, N., & Abramson, C. M. (2020). Conclusion: A comparative analysis of comparative ethnographies. In C. M. Abramson & N. Gong. (Eds.), *Beyond the case: The logics and practices of comparative ethnography* (pp. 283–308). Oxford University Press.
Grigoropoulou, N., & Small, M. L. (2022). The data revolution in social science needs qualitative research. *Nature Human Behaviour, 6*(7), 904–906.
Grimmer, J., Roberts, M. E., & Stewart, B. M. (2021). Machine learning for social science: An agnostic approach. *Annual Review of Political Science*, *24*(1), 395–419.
Grimmer, J., Roberts, M. E., & Stewart, B. M. (2022). *Text as data: A new framework for machine learning and the social sciences*. Princeton University Press.
Hajian, S., Bonchi, F., & Castillo, C. (2016). Algorithmic bias: From discrimination discovery to fairness-aware data mining. In B. Krishnapuram (Ed.), *Proceedings of the 22nd ACM SIGKDD International Conference on Knowledge Discovery and Data Mining* (pp. 2125– 2126). Association for Computing Machinery.
Hanna, A., Denton, E., Smart, A., & Smith-Loud, J. (2020). Towards a critical race methodology in algorithmic fairness. In *Proceedings of the 2020 Conference on Fairness, Accountability, and Transparency* (501–512). Association for Computing Machinery.
Hine, C. (2011). *Virtual ethnography*. Sage.
Hutson, M. (2021, March 3). Robo-writers: The rise and risks of language-generating AI. *Nature*. https://www.nature.com/articles/d41586-021-00530-0
Jackson, J. E. (2015). Changes in fieldnotes practice over the past thirty years in U.S. anthropology. In R. Sanjek & S. W. Tratner (Eds.), *eFieldnotes: The makings of anthropology in the digital world* (pp. 42–64). University of Pennsylvania Press.
Jordan, M. I., & Mitchell, T. M. (2015). Machine learning: Trends, perspectives, and prospects. *Science, 349*(6245), 255–260.
Knox, H., & Nafus, D. (2018). *Ethnography for a data-saturated world*. Manchester University Press.
Koenecke, A., Nam, A., Lake, E., Nudell, J., Quartey, M., Mengesha, Z., Toups, C., Rickford, J. R., Jurafsky, D., & Goel, S. (2020). Racial disparities in automated speech recognition. *Proceedings of the National Academy of Sciences of the United States of America*, *117*(14), 7684–7689.
Lamont, M., & Swidler, A. (2014). Methodological pluralism and the possibilities and limits of interviewing. *Qualitative Sociology*, *37*(2), 153–171.
Lange, A.-C., Lenglet, M., & Seyfert, R. (2019). On studying algorithms ethnographically: Making sense of objects of ignorance. *Organization*, *26*(4), 598–617.
Lara-Millán, A. (2021). *Redistributing the poor: Jails, hospitals, and the crisis of law and fiscal austerity*. Oxford University Press.
Lareau, A. (2003). *Unequal childhoods: Class, race, and family life*. University of California Press.
Lazer, D., Kennedy, R., King, G., & Vespignani, A. (2014). The parable of Google flu: Traps in big data analysis. *Science*, *343*(6176), 1203–1205.
Le, Q. V., & Mikolov, T. (2014). Distributed representations of sentences and documents. arXiv. Article 1405.4053. https://arxiv.org/pdf/1405.4053.pdf
Li, Z. (2022, December 8). ASA2022_Workshop Commits. GitHub. https://github.com/lizhuofan95/ASA2022_Workshop/blob/main/ASA_Working_003_20220720_PUBLIC.ipynb
Li, Z., Dohan, D., & Abramson, C. M. (2021). Qualitative coding in the computational era: A hybrid approach to improve reliability and reduce effort for coding ethnographic interviews. *Socius*, *7*, Article 23780231211062344.

Lichterman, P., & Reed, I. A. (2015). Theory and contrastive explanation in ethnography. *Sociological Methods & Research*, *44*(4), 585–635.
Lubet, S. (2018). *Interrogating ethnography: Why evidence matters*. Oxford University Press.
McFarland, D. A., & McFarland, R. H. (2015). Big data and the danger of being precisely inaccurate. *Big Data & Society*, *2*(2), Article 205395171560249.
Mikolov, T., Chen, K., Corrado, G., & Dean. J. (2013). Efficient estimation of word representations in vector space. arXiv. Article 1301.3781. https://arxiv.org/abs/1310.4546
Mimno, D. (n.d.). jsLDA: In-browser topic modeling [Computer software]. David Mimno. https://mimno.infosci.cornell.edu/jsLDA
Mohr, J. W. (1998). Measuring meaning structures. *Annual Review of Sociology*, *24*(1), 345–370.
Molina, M., & Garip, F. (2019). Machine learning for sociology. *Annual Review of Sociology*, *45*(1), 27–45.
Murphy, A. K., Jerolmack, C., & Smith, D. (2021). Ethnography, data transparency, and the information age. *Annual Review of Sociology*, *47*(1), 41–61.
Nelson, L. K. (2020). Computational grounded theory: A methodological framework. *Sociological Methods & Research*, *49*(1), 3–42.
Nelson, L. K., Burk, D., Knudsen, M., & McCall, L. (2021). The future of coding: A comparison of hand-coding and three types of computer-assisted text analysis methods. *Sociological Methods & Research*, *50*(1), 202–237.
Oliver, D. G., Serovich, J. M., & Mason, T. L. (2005). Constraints and opportunities with interview transcription: Towards reflection in qualitative research. *Social Forces*, *84*(2), 1273–1289. Padgett, J. F., Prajda, K., Rohr, B., & Schoots, J. (2020). Political discussion and debate in narrative time: The Florentine *Consulte e pratiche*, 1376–1378. *Poetics*, *78*, Article 101377.
Pardo-Guerra, J. P., & Pahwa. P. (2022). The extended computational case method: A frame- work for research design. *Sociological Methods & Research*, Article 00491241221122616.
Paszke, A., Gross, S., Massa, F., Lerer, A., Bradbury, J., Chanan, G., Killeen, T., Lin, Z., Gimelshein, N., Antiga, L., Desmaison, A., Köpf, A., Yang, E., DeVito, Z., Raison, M., Tejani, A., Chilamkurthy, S., Steiner, B., Fang, L., Bai, J., & Chintala, S. (2019). PyTorch: An imper- ative style, high-performance deep learning library. arXiv. Article 1912.01703. http://arxiv. org/abs/1912.01703
Reich, A. D. (2014). *Selling our souls: The commodification of hospital care in the United States*. Princeton University Press.
Rosa, F. R. (2022). Code ethnography and the materiality of power in internet governance. *Qualitative Sociology*, *45*(3), 433–455.
Rule, A., Cointet, J.-P., & Bearman, P. S. (2015). Lexical shifts, substantive changes, and continuity in State of the Union discourse, 1790–2014. *Proceedings of the National Academy of Sciences of the United States of America*, *112*(35), 10837–10844.
Sallaz, J. J. (2019). *Lives on the line*. Oxford University Press.
Sánchez-Jankowski, M. (2008). *Cracks in the pavement: Social change and resilience in poor neighborhoods*. University of California Press.
Seaver, N. (2017). Algorithms as culture: Some tactics for the ethnography of algorithmic systems. *Big Data & Society*, *4*(2), Article 2053951717738104.
Small, M. L. (2009). "How many cases do I need?': On science and the logic of case selection in field-based research. *Ethnography*, *10*(1), 5–38.
Szeliski, R. (2022). *Computer vision: Algorithms and applications* (2nd ed.). https://szeliski.org/ Book/.
Taylor, T. L. (2018). *Watch me play: Twitch and the rise of game live streaming*. Princeton University Press.
Timmermans, S., & Tavory, I. (2012). Theory construction in qualitative research: From grounded theory to abductive analysis. *Sociological Theory*, *30*(3), 167–186.
Van Noorden, R. (2022, April 28). How language-generation AIs could transform science. *Nature*. https://www.nature.com/articles/d41586-022-01191-3
Venkatesh, S. A. (2009). *American project: The rise and fall of a modern ghetto*. Harvard University Press.
Wang, T. (2012). Writing live fieldnotes: Towards a more open ethnography. *Ethnography Matters*. https://ethnographymatters.net/blog/2012/08/02/writing-live-fieldnotes-towards- a-more-open-ethnography/
Whyte, W. F. (2012). *Street corner society: The social structure of an Italian slum*. University of Chicago Press. (Original work published 1943).
Willis, P. E. (1981). *Learning to labor: How working class kids get working class jobs*. Columbia

University Press. (Original work published 1977).
Wilson, S. M., & Peterson, L. C. (2002). The anthropology of online communities. *Annual Review of Anthropology*, *31*(1), 449–467.
Wolf, T., Debut, L., Sanh, V., Chaumond, J., Delangue, C., Moi, A., Cistac, P., Rault, T., Louf, R., Funtowicz, M., Davison, J., Shleifer, S., von Platen, P., Ma, C., Jernite, Y., Plu, J., Xu, C., Le Scao, T., Gugger, S., Drame, M., Lhonest Q., Rush, A. (2020). Transformers: State-of-the-Art Natural Language Processing. *Proceedings of the 2020 Conference on Empirical Methods in Natural Language Processing: System Demonstrations*, 38–45.